\documentclass[runningheads]{llncs}
\usepackage{graphicx}
\usepackage{amsmath}
\usepackage{amsfonts}
\usepackage{multirow}
\usepackage{booktabs}   
\usepackage{pifont}
\usepackage{xcolor}
\usepackage{hyperref}
\usepackage{tabularx}
\hypersetup{colorlinks,linkcolor={blue},citecolor={green},urlcolor={red}} 
\usepackage{hhline}

\newcommand{\cmark}{\ding{51}}%

\definecolor{ForestGreen}{RGB}{34,139,34}
\newcommand\improve[1]{\textcolor{ForestGreen}{\textbf{#1}}}

\newcommand{\sag}[0]{\texttt{SAG}}
\begin{document}
\title{Semantics-Aware Attention Guidance for Diagnosing Whole Slide Images}
%
%

\author{Kechun Liu \inst{1}$^{,}$\thanks{contribute equally to this work.} \and
        Wenjun Wu \inst{1}$^{,\star}$ \and
        Joann G. Elmore\inst{2} \and
        Linda G. Shapiro\inst{1}}

\authorrunning{Liu et al.}  

\institute{University of Washington, Seattle, WA, 98195, USA \and
            David Geffen School of Medicine, UCLA, Los Angeles, CA 90024, USA\\
\email{kechun@cs.wasghinton.edu}}

\maketitle              
\begin{abstract}
Accurate cancer diagnosis remains a critical challenge in digital pathology, largely due to the gigapixel size and complex spatial relationships present in whole slide images. Traditional multiple instance learning (MIL) methods often struggle with these intricacies, especially in preserving the necessary context for accurate diagnosis. In response, we introduce a novel framework named Semantics-Aware Attention Guidance (\sag{}), which includes 1) a technique for converting diagnostically relevant entities into attention signals, and 2) a flexible attention loss that efficiently integrates various semantically significant information, such as tissue anatomy and cancerous regions. Our experiments on two distinct cancer datasets demonstrate consistent improvements in accuracy, precision, and recall with two state-of-the-art baseline models. Qualitative analysis further reveals that the incorporation of heuristic guidance enables the model to focus on regions critical for diagnosis. \sag{} is not only effective for the models discussed here, but its adaptability extends to any attention-based diagnostic model. This opens up exciting possibilities for further improving the accuracy and efficiency of cancer diagnostics. Upon acceptance, our code will be made available. 

\end{abstract}

\section{Introduction}

In recent years, the landscape of histopathological image analysis has been profoundly reshaped by the advent of deep learning technologies \cite{echle2021deep,chen2022recent}. However, learning from gigapixel whole slide images (WSIs) remains a difficult problem, as their size makes end-to-end learning extremely expensive. Thus, WSI classification methods often follow a bag-of-words (BoW) model for learning representations, wherein a large patch of a whole slide image is treated as a bag or set, while smaller image patches inside a bag are treated as words (or instances). Following this BoW model, many studies adopt a multiple instance learning-based (MIL) approach, which involves first extracting word-level feature representations and then applying global aggregation to bags of word-level representations to obtain WSI-level representations. These approaches are good at reducing the computational cost and offer a workaround by segmenting WSIs into smaller, and more manageable patches \cite{hou2016patch,ilse2018attention,mercan2020deep,li2021dual}.

However, how pathologists approach diagnosis is very different from MIL models.  Pathologists begin their evaluation by identifying suspicious regions at low magnification to form initial hypotheses. They then switch to high magnification to examine individual cells, mitotic counts, structures like ducts, and etc., ultimately reaching a definitive diagnosis~\cite{mello2012perceptual}. In contrast, by treating image patches independently, MIL models disregard the multi-scale nature of pathology, where zooming in and out is crucial for comprehensive assessment. This limitation in capturing long-range interactions between entities hinders MIL models from effectively capturing the nuanced details critical for accurate diagnosis. \par

\begin{figure*}[t!]
    \centering
    \includegraphics[width=1.0\columnwidth]{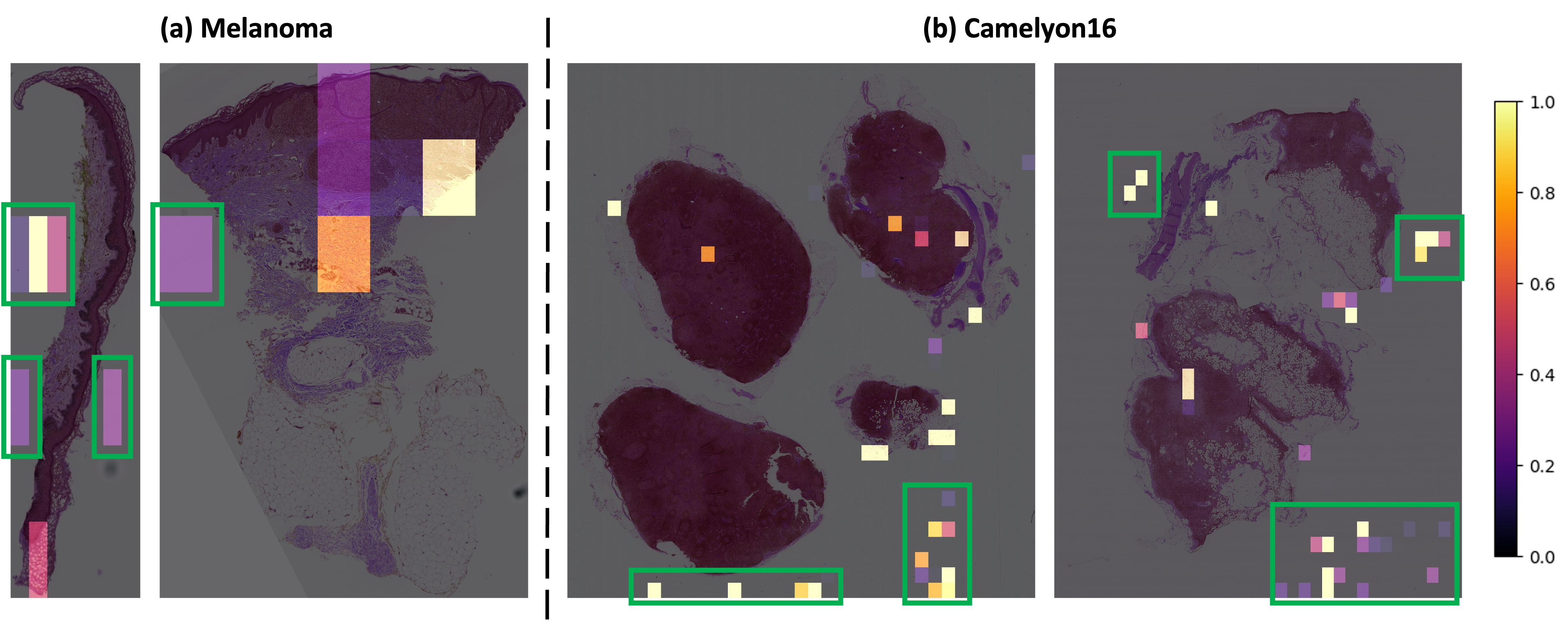}
    \caption{Visualization of the baseline model's (ScAtNet \cite{wu2021scale}) attention on (a) skin biopsy WSIs in the melanoma dataset and (b) breast biopsy WSIs in the Camelyon16 dataset. Green boxes show examples of the baseline model mistakenly focusing on background regions. The signal and attention values are normalized for visualization purposes.} 
    \label{fig:motivation}
\end{figure*}

To learn a better global representation, transformer models have been adopted to grasp the interdependencies among patches and formulate comprehensive representations, notably advancing beyond the MIL's limitations \cite{myronenko2021accounting,chen2021multimodal,chen2022scaling,zheng2022graph,wu2021scale}. A few studies extract features from multiple resolutions and aggregate them hierarchically or concatenate them to predict the diagnosis class \cite{wu2021scale,guo2023higt,shi2023structure}. Specifically, ScAtNet \cite{wu2021scale} employs a transformer-based end-to-end network that adapts to the information from different input scales through self-attention and predicts the classification label. Results show that ScAtNet outperforms other MIL methods by a large margin in the task of melanoma diagnosis. However, such models often mistakenly focus on non-cancerous regions or just empty spaces, as highlighted by the green boxes in Fig.~\ref{fig:motivation}. This problem brings up questions about how well these models can be interpreted, how reliable they are, and if they really match up with the way pathologists diagnose. 
\par
In response, integrating additional domain information into diagnostic models has emerged as a promising strategy. Such efforts not only enhance classification accuracy but also improve model performance, especially in scenarios where data is scarce. Miao \textit{et al.} introduce spatial prior attention using binary anatomy knowledge maps, a step towards integrating prior knowledge into WSI diagnosis \cite{miao2022prior}. Limited to binary representations, this study suggests the potential for richer prior knowledge to improve accuracy. Chen \textit{et al.} broadens the scope by leveraging genomics information in addition to WSIs to predict patient outcomes~\cite{chen2021multimodal}. Yet, their approach lacks adaptability to other modalities and is hard to integrate additional guidance signals. \par
Recognizing the limitations of current methods, we propose a Semantics-Attention-Guiding framework, \sag{}, whose key contributions are:
\begin{itemize}
    \item A novel attention guiding module that is applicable to any attention-based multiple instance learning or Transformer models. 
    \item A flexible attention-guiding loss to effectively incorporate varied semantic information, such as tissue and cancerous region masks.
    \item A heuristic attention-generation method to convert diagnostically relevant entities to heuristic-guidance signals.  
    \item Improving state-of-the-art methods on two datasets of different cancer types.
\end{itemize}


\section{Methodology}
\begin{figure*}[t!]
    \centering
    \includegraphics[width=1.0\columnwidth]{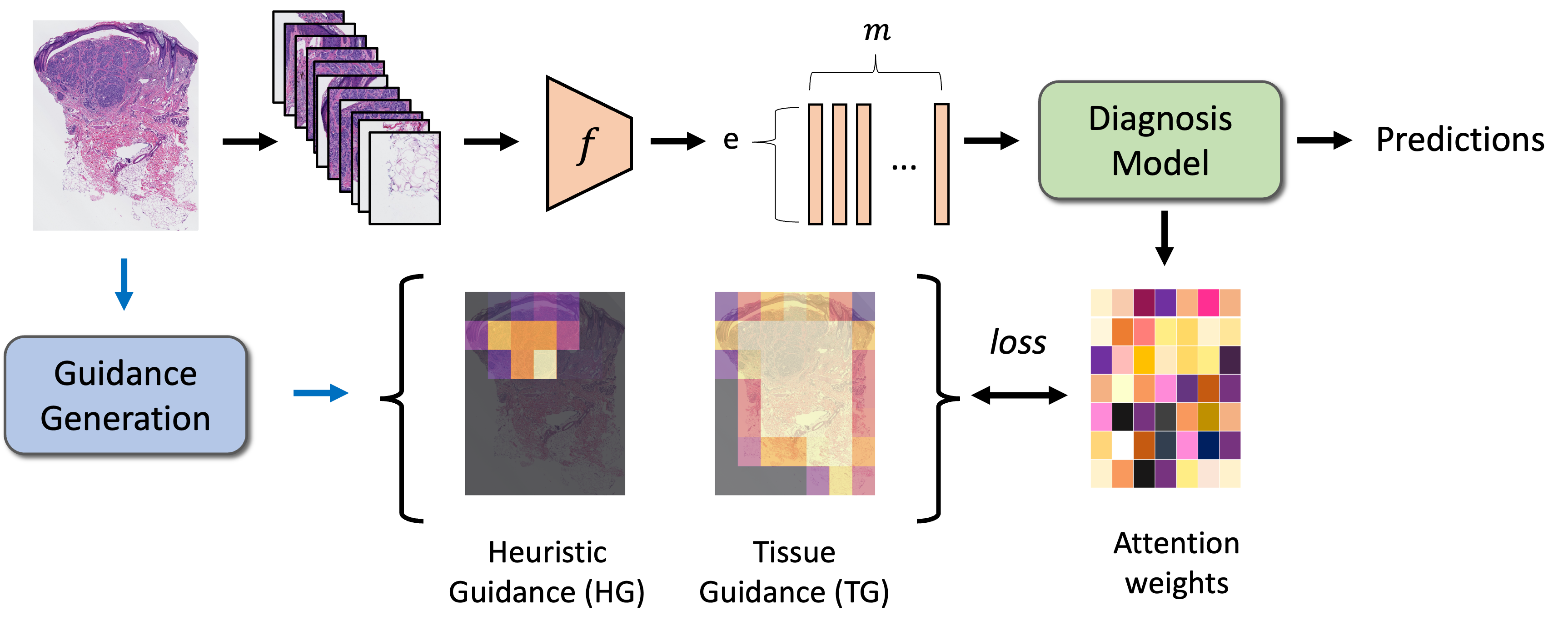}
    \caption{Overview of the \sag{} approach for improving WSIs diagnosis models. First, a high-resolution histopathological image is divided into $p$ number of non-overlapping patches. Then, patch embeddings are obtained using an off-the-shelf feature extractor \textit{f}. Subsequently, a diagnostic network utilizes the $p \times e$-dimensional feature map for classification into distinct categories. During training, heuristic guidance ($\mathbf{HG}$) and tissue guidance ($\mathbf{TG}$) are leveraged to supervise the attention within the diagnosis model, ensuring the focus on diagnostically relevant regions.} 
    \label{fig:pipeline}
\end{figure*}

Our \sag{} framework aims to infuse diagnostic models with relevant knowledge, thereby enhancing the diagnostic performance and the interpretability of attention-supervised representations. This versatile framework is compatible with a broad range of attention-based MIL and transformer methods. Fig.~\ref{fig:pipeline} illustrates our \sag{} pipeline, which includes three main components: 1) generate patchwise embeddings with an off-the-shelf feature extractor, 2) learn diagnostic patterns from these embeddings via a diagnosis network, and 3) utilize an attention-guiding loss that leverages heuristic guidance ($\mathbf{HG}$) and tissue guidance ($\mathbf{TG}$). In the following sections, we give the details of the proposed attention guidance.

\subsection{Diagnosis Models}
\label{sec:diagnosis_model}
We employ a pre-trained feature extractor $f$ for patch embedding extraction. The implementation detail of $f$ is provided in Sec.~\ref{sec:implementation}. Moreover, to demonstrate the versatility and model-agnostic nature of our \sag{} framework, we apply \sag{} to two state-of-the-art baseline models: a transformer-based model, ScAtNet \cite{wu2021scale}, and an MIL-based model, ABMIL \cite{ilse2018attention}.

\subsection{Attention Weights}
First, we partition an image into $p$ input patches. For transformer-based models, the architecture consists of $l$ layers with $h$ self-attention heads per layer. Given embeddings $q, k, v \in \mathbb{R}^{p\times d_k}$ projected from the inputs, each attention head induces a pairwise similarity from query $q$ and key $k$ to transform the value $v$. The similarity ($\mathbf{A}$) and the model attention weights ($\mathbf{MA_t}$) of the transformers are computed as follows:
\begin{equation}
    \begin{aligned}
    \label{eqn:attn_weights}
    &\mathbf{A} = \textnormal{softmax}(\frac{qk^\top}{\sqrt{D_h}}) \in \mathbb{R}^{p\times p}, \\
    &\mathbf{MA_t} = \frac{1}{p}\sum_{i=1}^p A_i \in \mathbb{R}^{p}.
    \end{aligned}
\end{equation}

The model attention weights ($\mathbf{MA_m}$) of the MIL methods are formulated as the weighted aggregation of instance embeddings~\cite{ilse2018attention}:
\begin{equation}
    \mathbf{MA_m} = \sigma(x) \in \mathbb{R} ^ {p},
\end{equation}
where $\sigma$ denotes the linear layers to learn the attention weights, and $x\in \mathbb{R}^{p\times d}$ denotes the embeddings from $p$ patches.

\subsection{Guidance Generation}
\label{sec:HG}
\begin{figure*}[t!]
    \centering
    \includegraphics[width=1.0\columnwidth]{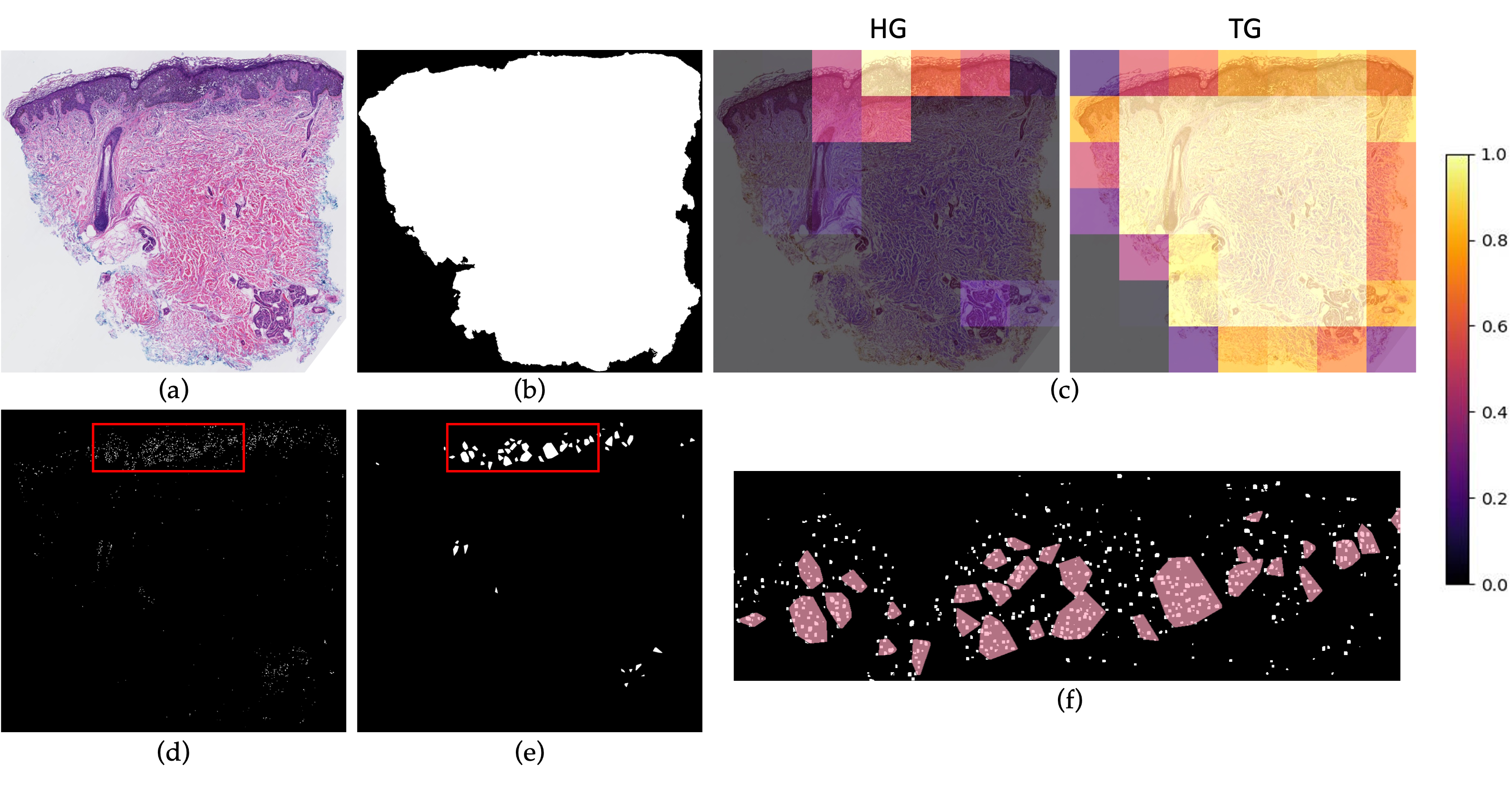}
    \caption{Generation of attention guidance: (a) H\&E sample image. (b) Tissue segmentation mask. (c) $\mathbf{HG}$ and $\mathbf{TG}$. The values are normalized for visualization purpose. (d) Cellular entities detected (\textbf{zoom-in for best view}). (e) Convex hull of cellular clusters. (f) A zoomed-in view of the red boxes in (d) and (e). The convex hull is rendered with red color.} 
    \label{fig:attention_guidance}
\end{figure*}

To regularize the model's attention $\mathbf{MA}$, we induce two types of semantic attention guidance: tissue guidance ($\mathbf{TG}$) and heuristic guidance ($\mathbf{HG}$) (Fig.~\ref{fig:attention_guidance}), each represented as a vector $\in \mathbb{R}^p$. The generation of attention guidance is described in two steps: 1) Acquisition of tissue mask and diagnostic heuristics, and 2) Calculation of guidance weights.

To obtain the tissue mask for $\mathbf{TG}$, Otsu's method \cite{zhang2008image} is used to perform high-quality segmentation of tissue patches. This process transforms the input image shown in Fig.~\ref{fig:attention_guidance}a into the binary tissue mask shown in Fig.~\ref{fig:attention_guidance}b. \par
To obtain $\mathbf{HG}$, we exploit dataset- and disease-specific prior knowledge, such as structures, tissues, and cells. In the example shown in Fig.~\ref{fig:attention_guidance}, we first perform cell segmentation for a specific cell type (Fig.~\ref{fig:attention_guidance}d). Then, groups of cells are aggregated via the density-based spatial clustering algorithm DBSCAN~\cite{ester1996density}. Next, the convex hull~\cite{2020SciPy-NMeth} is generated for each cluster (Fig.~\ref{fig:attention_guidance}e) and utilized as the semantic signal for attention supervision (Fig.~\ref{fig:attention_guidance}f).

To calculate the guidance weight $W \in \mathbb{R}^p$, we leverage Eqn.~\ref{eq:guidance} to transform the heuristic signals (\textbf{HG}) and the tissue masks (\textbf{TG}) into the attention supervision (Fig.~\ref{fig:attention_guidance}c):
\begin{equation}
\label{eq:guidance}
    W_i^k = \frac{M_i^k}{\sum_{j=1}^{p} M_j^k}, \quad k\in\{\mathbf{TG}, \mathbf{HG}\},
\end{equation}
where $W_i^k$ denotes the guidance weight of patch $i$, and $M_i^k$ is the mask area ratio of patch $i$.

\subsection{Loss Functions}
Since heuristic guidance ($\mathbf{HG}$) reflects the relevance to the diagnosis, we employ the mean squared error (MSE) loss, $L_{mse}$, to regularize $\mathbf{MA}$: 
\begin{equation}
    L_{mse} = \frac{1}{p}\sum_{i=1}^p (W_i^\mathbf{HG} - \mathbf{MA}_i)^2.
\end{equation}

On the other hand, tissue guidance ($\mathbf{TG}$) is useful in guiding the model to focus on tissue patches and ignore the background and artifact patches. Thus, we employ a less constrained loss, $L_{in\&out}$, which sums the attention weights outside of the tissue and the negative attention weights inside the tissue, as defined in Eqn.~\ref{eqn:inout} below:
\begin{equation}
\label{eqn:inout}
    L_{in\&out} = \frac{1}{p}(-\sum_{i, W_i^{\mathbf{TG}}>0}^p\textbf{MA}_i +  \sum_{i, W_i^{\mathbf{TG}}=0}^p\textbf{MA}_i).
\end{equation}

For joint learning, we leverage uncertainty weighting, $\mathcal{UW}$~\cite{kendall2018multi}, which weighs multiple loss functions by considering the homoscedastic uncertainty of each task. The overall loss function is defined as:
\begin{equation}
    L = \mathcal{UW}\otimes\{L_{cls}, L_{mse}, L_{in\&out}\},
\end{equation}
where $L_{cls}$ is the cross entropy loss for the classification task. 
\section{Experiments and Results}

\subsection{Datasets}
\textbf{Melanoma.}
The melanoma diagnosis dataset used in the study consists of 222 H\&E stained WSIs. There are four classes in this dataset: 1) mild and moderate dysplastic nevi, 2) melanoma in situ, 3) invasive melanoma stage pT1a, and 4) invasive mealnoma stage $\geq$ pT1b. In our study, we use a random split of 89/22/111 samples for training, validation and testing. We follow the preprocessing steps in ScAtNet\cite{wu2021scale} which crops the slice into 25, 49, and 81 patches in 7.5x, 10x, and 12.5x magnifications. \par

\noindent
\textbf{Camelyon16.}
Camelyon16~\cite{bejnordi2017diagnostic} is a public dataset comprising 400 H\&E stained WSIs from breast cancer. The WSIs are diagnosed into two classes: normal and tumor. We use the official split of 271/129 slides for training and testing. To train ABMIL, we follow DSMIL~\cite{li2021dual}, which crops the WSI into 224x224 sized non-overlapping patches in 20x magnification, and excludes background patches, leaving around 15K patches per bag on average. To train ScAtNet on breast biopsies, we adapted the original skin biopsy patch size while adjusting the number of patches per WSI (10x magnification) to maintain similar content per patch. The result is $35 \times 35$, or $1,225$ number of crops. This ensures consistent representation and preserves model architecture.

\subsection{Implementation Details}
\label{sec:implementation}
\textbf{Feature Extraction and Attention Guidance.}
For the melanoma dataset, an ImageNet pre-trained MobileNetV2~\cite{sandler2018mobilenetv2} extracts a 1280-dimensional feature vector for each patch, as described in Sec.~\ref{sec:diagnosis_model}. Since melanocytes are believed to be highly informative about melanoma diangosis, an open-sourced off-the-shelf melanocyte detection model~\cite{liu2023vsgd} is employed to generate the cellular entity map that eventually transforms to $\mathbf{HG}$, as described in Sec.~\ref{sec:HG}.  To cluster the cell entities, DBSCAN in the scikit-learn package \cite{sklearn_api} is used with \texttt{eps=20} and \texttt{min\_samples=5}. $\mathbf{TG}$ is generated using Otsu thresholding \cite{zhang2008image}.

For Camelyon16~\cite{bejnordi2017diagnostic}, a SimCLR pretrained by DSMIL~\cite{li2021dual} extracts a 512-dimensional feature vector for each patch. Moreover, the metastasis mask and tissue mask in the dataset are utilized for $\mathbf{HG}$ and $\mathbf{TG}$.

\noindent
\textbf{Diagnosis Models and Training Details.}  \sag{} is applied to two models: a transformer model, ScAtNet~\cite{wu2021scale}, and a MIL model, ABMIL~\cite{ilse2018attention}. For ScAtNet, we impose $\mathbf{TG}$ across all attention heads and impose $\mathbf{HG}$ on half of the attention heads. This maintains the model's adaptability and accommodates potential noise in $\mathbf{HG}$. For ABMIL, we apply both $\mathbf{HG}$ and $\mathbf{TG}$ on the melanoma dataset, while we only apply $\mathbf{HG}$ to Camelyon16 as the dataset already exclude background patches. We use ABMIL's~\cite{ilse2018attention} and ScAtNet's~\cite{wu2021scale} public codebase for implementation and train models under their experimental settings. 

\subsection{Results}
Table~\ref{tab:results} compares the overall performance of \sag{} on different datasets and backbone models, demonstrating its consistent ability to enhance diagnostic performance in histopathological image analysis. For each setting, we conduct $15$ runs of experiments with randomly sampled seeds and report the average. \par

\begin{table}[h!]
\centering
\caption{Experimental Results of SAG across single-scale (SC) and multi-scale (MC) configurations
for Melanoma and Camelyon16 datasets. Baseline methods are indicated with a $\dagger$. Performance metrics include Accuracy (Acc), Precision (P), Recall (R), and Area Under the Curve (AUC).}

\label{tab:results} 
\scriptsize
\begin{tabularx}{\linewidth}{l|cc|XXXX|XXXX}
\toprule
 & \multicolumn{2}{c|}{\sag{}} & \multicolumn{4}{c|}{Melanoma} & \multicolumn{4}{c}{Camelyon16} \\
\midrule
Methods & HG &  TG & Acc & P & R & AUC & Acc & P & R & AUC \\
\hline
ScAtNet (SC)$\dagger$\cite{wu2021scale} & & & 55.03 & 57.17 & 55.36 & 77.38 & 67.79 & 58.17 & 57.51 & 70.28 \\
ScAtNet (SC) & \cmark & & 57.14 & 59.57 & 57.31 & 78.75 & 68.71 & 58.50 & 64.01 & 72.39 \\
ScAtNet (SC) & \cmark & \cmark & 56.67 & 60.27 & 56.66 & 79.72 & \textbf{71.60} & \textbf{64.45} & 61.22 & 71.87 \\
ScAtNet (MC)$\dagger$ & & & 58.16 & 61.54 & 58.21 & 79.54 & 66.82 & 55.98 & 61.22 & 69.45\\
ScAtNet (MC) & \cmark & & 59.95 & 64.77 & 60.13 & 81.58 & 67.91 & 57.28 & \textbf{66.39} & 72.26 \\
ScAtNet (MC) & \cmark & \cmark & \textbf{62.71} & \textbf{65.23} & \textbf{63.34} & \textbf{82.03} & 70.13 & 60.53 & 62.58 & \textbf{73.13} \\
\hline
\addlinespace[3pt]
Best Improvement $\triangle$ &  & & \improve{+4.55} & \improve{+3.69} & \improve{+5.13} & \improve{+2.49} & \improve{+3.81} & \improve{+6.28} & \improve{+6.50} & \improve{+3.68}\\
\addlinespace[3pt]
\hline
\hline
\addlinespace[3pt]
ABMIL$\dagger$\cite{ilse2018attention} & & & 45.55 & 48.23 & 46.42 & 68.07 & 93.02 & 92.47 & 92.79 & 97.52 \\
ABMIL & \cmark & & 51.59 & \textbf{57.42} & 51.02 & \textbf{74.68} & \textbf{94.73} & \textbf{94.61} & \textbf{94.17} & \textbf{97.80}\\
ABMIL & \cmark & \cmark & \textbf{52.01} & 56.25 & \textbf{51.84} & 74.35 &\multicolumn{4}{c}{\textit{Not Applicable}}\\
\hline
\addlinespace[3pt]
Best Improvement $\triangle$& & & \improve{+6.46} & \improve{+9.19} & \improve{+5.42} & \improve{+6.28} & \improve{+1.71} & \improve{+2.14} & \improve{+1.38} & \improve{+0.28} \\
\bottomrule
\end{tabularx}
\end{table}

Notably, incorporating \sag{} into single- and multi-scale ScAtNet models on the melanoma dataset yields significant improvements, particularly with multi-scale inputs achieving a $4.55\%$ accuracy increase (Table~\ref{tab:results}). Similar trends are observed on Camelyon16, where \sag{} boosts accuracy across ScAtNet configurations ($3.81\%$ for multi-scale) and increases ABMIL's accuracy by $1.71\%$ (Table~\ref{tab:results}). These improvements highlight \sag{}'s effectiveness in refining focus and enhancing the models' diagnostic performance.\par
In our analysis, we observe that ABMIL exhibits superior diagnostic performance on the Camelyon16 dataset ($94.73\%$ vs. $71.60\%$), whereas ScAtNet is more effective on the melanoma dataset ($62.71\%$ vs $45.52\%$). This distinction in model efficacy can be attributed to the intrinsic characteristics of these datasets and the models' specific designs. Notably, our melanoma dataset, presenting a four-class classification problem, requires a comprehensive understanding of the entire image at multiple scales and holistic levels. This aligns well with ScAtNet's transformer-based architecture, which excels at capturing long-range dependencies and aggregating multi-scale information through attention mechanisms \cite{wu2021scale}. In contrast, the Camelyon16 dataset, being a binary classification problem, prioritizes local feature identification for diagnosis, which aligns with ABMIL's MIL-based approach, suggesting why ABMIL outperforms in this context. On the other hand, ScAtNet's complexity and multi-scale inputs may not offer significant benefits here due to overfitting risks. This highlights the importance of choosing an appropriate method based on the specific data characteristics. \par

\begin{figure*}[t!]
    \centering
    \includegraphics[width=1.0\columnwidth]{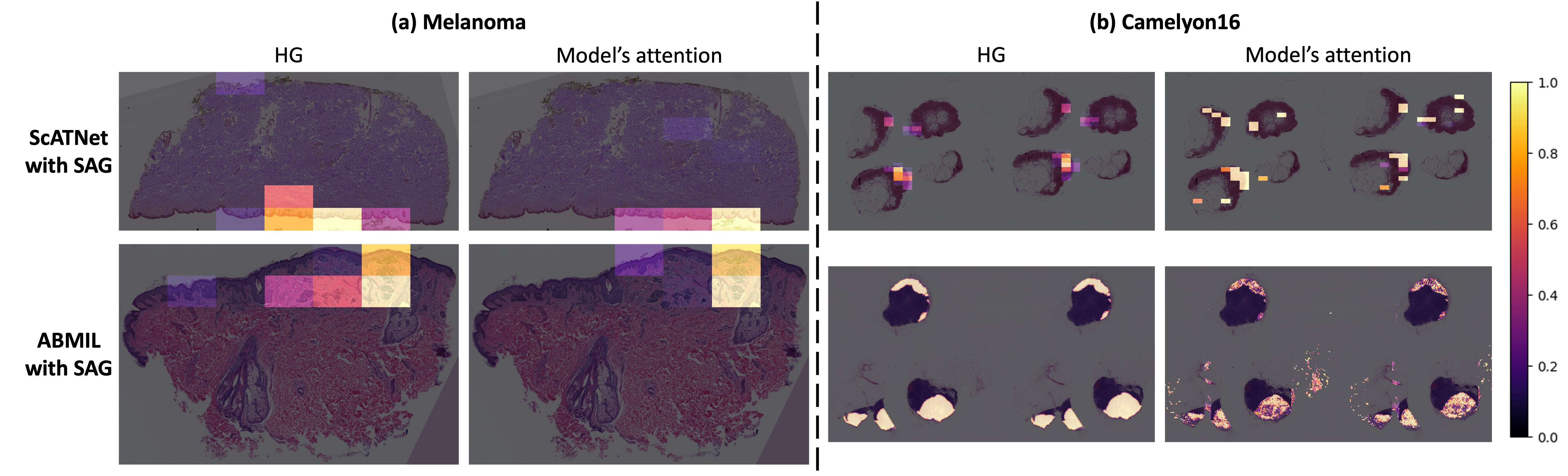}
    \caption{Comparative visualizations of $\mathbf{HG}$ and the models' attention under \sag{}'s training on the melanoma and Camelyon16 datasets. The images are sampled from test set. The $\mathbf{HG}$ and attention values are normalized for visualization purpose.} 
    \label{fig:results}
\end{figure*}

To further illustrate, Fig.~\ref{fig:results} visualizes the attention patterns of ScAtNet and ABMIL on both datasets compared to $\mathbf{HG}$. We notice that \sag{} encourages the model to focus on diagnostically relevant regions. These visualizations effectively demonstrate \sag{}'s capacity to guide attention and improve interpretability. Additional visualizations are available in the appendix for further exploration.



\section{Conclusion}
Motivated by our observation of misplaced attention on irrelevant regions in previous approaches, we propose a novel framework called Semantics-Aware Attention Guidance (\sag{}). \sag{} integrates tissue and heuristic attention guidance to better emulate the diagnostic process of pathologists, focusing on meaningful interconnections within WSIs. This targeted approach enables \sag{} to enhance model performance across various datasets with limited size and potentially noisy annotations, highlighting its contribution to improving the precision and reliability of computational diagnostics. 
\newpage
%
%
%
%
\bibliographystyle{splncs04}

\bibliography{ref.bib}
\end{document}


\title{Supplementary Material}
%
\author{Kechun Liu \inst{1}$^{,}$\thanks{contribute equally to this work.} \and
        Wenjun Wu \inst{1}$^{,\star}$ \and
        Joann G. Elmore\inst{2} \and
        Linda G. Shapiro\inst{1}}

\authorrunning{Liu et al.}  

\institute{University of Washington, Seattle, WA, 98195, USA \and
            David Geffen School of Medicine, UCLA, Los Angeles, CA 90024, USA\\
\email{kechun@cs.wasghinton.edu}}

\maketitle


\begin{figure*}[h]
    \centering
    \includegraphics[width=0.95\columnwidth]{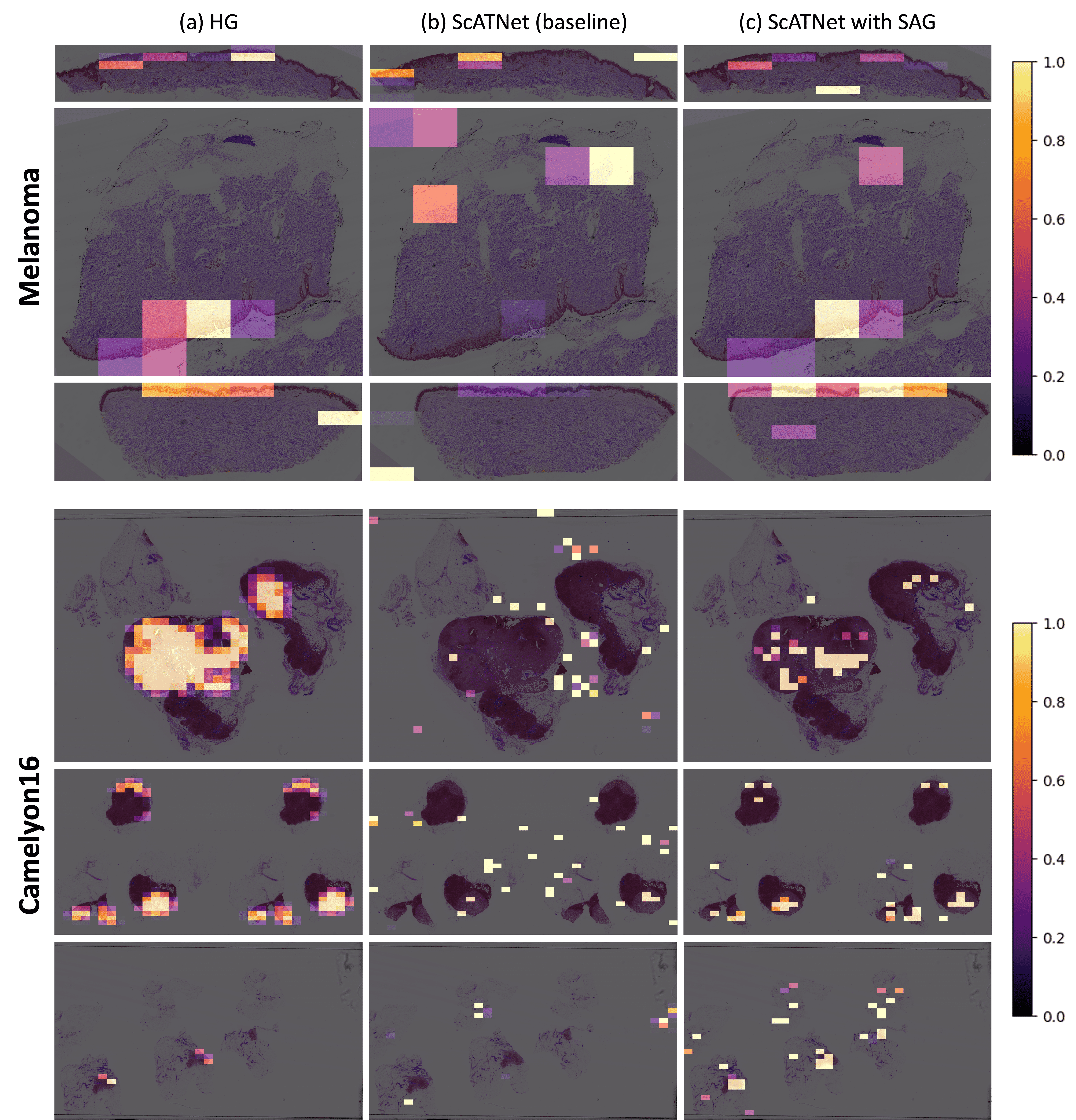}
    \caption{Comparison of (a) heuristic guidance ($\mathbf{HG}$), (b) ScAtNet (baseline)'s attention, and (c) ScAtNet (with \texttt{SAG})'s attention on the melanoma and Camelyon16 dataset. These images are sampled from the test set. The signal and attention weights are normalized for visualization purporse.} 
\end{figure*}

\begin{figure*}[h]
    \centering
    \includegraphics[width=0.95\columnwidth]{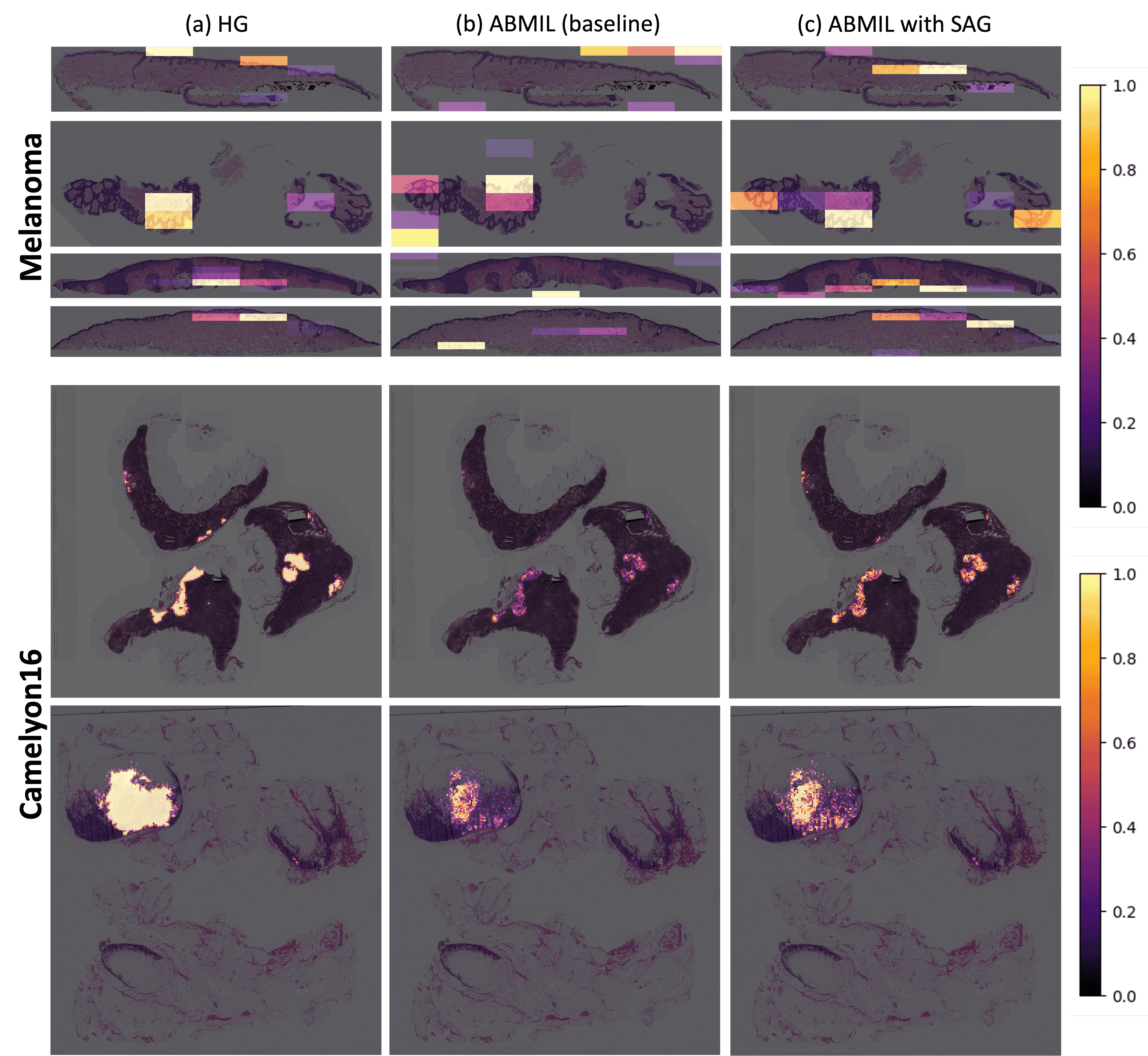}
    \caption{Comparison of (a) heuristic guidance ($\mathbf{HG}$), (b) ABMIL (baseline)'s attention, and (c) ABMIL (with \texttt{SAG})'s attention on the melanoma and the Camelyon16 dataset. These images are sampled from the test set. The signal and attention weights are normalized for visualization purporse.} 
\end{figure*}

